\documentclass{article}

\usepackage[utf8]{inputenc} 
\usepackage{url}            
\usepackage{booktabs}       
\usepackage{amsfonts}       
\usepackage{nicefrac}       
\usepackage{microtype}      
\usepackage{color}      

\usepackage{multirow}

\usepackage{comment}
\usepackage{float}
\usepackage{bm,amsmath,amsfonts}
\usepackage[inline]{enumitem}
\usepackage{subcaption}
\usepackage{mathtools}
\usepackage[a4paper]{geometry}
\usepackage[style=numeric,sorting=none]{biblatex}

\usepackage{amsthm}
\usepackage{amssymb}
\usepackage{amsmath,mathtools}
\usepackage{color}
\usepackage{caption}
\usepackage{algpseudocode}
\usepackage{algorithm}
\usepackage{indentfirst}
\usepackage{float}
\usepackage{breqn}
\usepackage[export]{adjustbox}

\newtheorem{theorem}{Theorem}[section]
\newtheorem{lemma}[theorem]{Lemma}

\newtheorem{proposition}[theorem]{Proposition}

\usepackage{amsmath}
\DeclareMathOperator*{\argmax}{arg\,max}

\newtheorem{problem}[theorem]{Problem}
\usepackage{biblatex}

\bibliography{bible.bib}
\begin{document}

\title{Reachability-based safe learning for optimal control problem}

\author{Stanislav Fedorov$^{\rm a}$\footnote{Corresponding author. e-mail: stanislav.fedorov@mathmods.eu}, Antonio Candelieri$^{\rm a}$ \\ \\       
        $^{\rm a}$\textit{Department of Computer Science, Systems and Communication, }\\
        \textit{University of Milano Bicocca, viale Sarca 336, Milan, 20126, Italy} \\
}

\maketitle
\vspace{5mm}

\begin{abstract}
	
	In this work we seek for an approach to integrate safety in the learning process that relies on a partly known state-space model of the system and regards the unknown dynamics as an additive bounded disturbance. We introduce a framework for safely learning a control strategy 
	for a given system with an additive disturbance. On the basis 
	of the known part of the model, a safe set in which the system can learn safely, 
	the algorithm can choose optimal actions for pursuing the target set as long as the safety-preserving 
	condition is satisfied. After some learning episodes, the disturbance can be updated based on real-world data. 
	To this end, Gaussian Process regression is conducted on the collected disturbance samples. 
	Since the unstable nature of the law of the real world, for example,  change of friction or conductivity with the temperature, we expect to have the more robust solution of optimal control problem.
	
	For evaluation of approach described above we choose an inverted pendulum as a benchmark model. The proposed algorithm manages to learn a policy that does not violate the pre-specified safety constraints. Observed performance is improved when it was incorporated exploration set up to make sure that an optimal policy is learned everywhere in the safe set. Finally, we outline some promising directions for future research beyond the scope of this paper.
	
\end{abstract}

\setcounter{page}{1}


\newcommand{\keyword}[1]{\textbf{#1}}
\newcommand{\tabhead}[1]{\textbf{#1}}
\newcommand{\code}[1]{\texttt{#1}}
\newcommand{\file}[1]{\texttt{\bfseries#1}}
\newcommand{\option}[1]{\texttt{\itshape#1}}

\section{Introduction}

Learning to control an uncertain system is a problem with many applications in various engineering fields. This work has been inspired by problems in autonomous and semi-autonomous systems. In the majority of practical scenarios, one wishes that the learning process terminates quickly and does not violate safety limits on key variables. It has to learn the control policy directly from experiments since there is no need to first derive an accurate physical model of the system. The main challenge when using such an approach is to ensure safety constraints during the learning process. However, constraining the freedom of the system can negatively affect performance, while attempting to learn less conservative safety constraints might fail to preserve safety if the learned constraints are inaccurate. Following  this problem, in this work we seek for an approach to integrate safety in the learning process that relies on a partly known state-space model of the system and regards the unknown dynamics as an additive bounded disturbance.

During related work for this paper has been reviewed different approaches for model to control modern Cyber-Physical Systems in the setting described above. We start with the general learning to control approach \cite{Goodfellow-et-al-2016} and concentrate on Bayesian sequential model-based optimization \cite{SeqOptBayes} with the purpose of defining technics for additive disturbance estimation. That also gives an idea of possible implementation of exploration step into the algorithm.

Generally, most of the works in safety during model optimization are based on celebrated Hamilton-Jacobi-Isaacs (HJI) reachability analysis, as for example \cite{conf/cdc/AkametaluKFZGT14},\cite{Ding_towardreachability-based}, which are taken as the basic papers for the future work. In \cite{conf/cdc/AkametaluKFZGT14} constraints bound made through Bayesian learning principles, more precisely has been used Gaussian Process (GP) model over the state space, which allows to model addition disturbances with infinite-dimensional basis of functions. This is done by choose to model the covariance function as a squared exponential distance between the states, the proof can be found for example in \cite{konig2013eigenvalue}. 
By considering worst-case disturbances, this method determines a safe region in the state space and provides a control policy to stay within that region. The main advantage is that in the interior of this region one can execute any desired action as long as the safe control is applied at the boundary, leading to a least restrictive control law. The desired action can be specified by any method, including any
learning algorithm. 

But the proposed technic is not without shortcomings, as HJI reachability analysis is done through solving PDE, which involves huge computation capacity. Then, in order to guarantee safety, the system designer must often rely on a nominal model that assumes conservative worst-case disturbances, which reduces the computed safe region, and thereby the region where the system can learn. Moreover, the least restrictive control law framework decouples safety and learning, which can lead to poor results, since it assumes drastic changes of control policy, due to discrete type of the response on safety constraints, and can lead to unexpected effects in the physics of the system as well as the learning controller has no notion of the unsafe regions of the state space and may attempt to drive the system into them.  Has been  reviewed the different approaches in the direction of reachability analysis, and one used in \cite{abate2008probabilistic} has been chosen as a basic, with further extensions. As a benchmark for safe learning approach has been chosen an inverted pendulum as a classic model. This 
system has the advantage to only have two states so that it can be analyzed simply and accurate, and the results can be illustrated fairly easily.

\section{Methodological background}

\subsection{Problem Setup}

For supervised learning it is needed a preliminary information. Assume that we have a training set $\mathcal{D}$  of $n$ observations,  $\mathcal{D} = \{ x_i,y_i \mid i = 1, \ldots, n \}$, where $x$ denotes an input vector of dimension $D$ and $y$ denotes a scalar output or target dependent variable; the column vector inputs for the design matrix all $n$ cases are aggregated in the $D \times n$ design matrix $X$, and the observed values are collected in the vector $y$, so we can write $\mathcal{D} = (X, y) $. Additionally, assume that we have a surrogate model, for which we partly know the dynamics of the system, and it is affected by additive i.i.d. noise $y = f(x) + \varepsilon$, where $\varepsilon \sim \ \mathcal{N} (0,\sigma_n^2)$. For model this phenomena, consider the following non-linear stochastic discrete time dynamical control system:

\begin{equation}\label{dynamics1}
x_{k+1} = f(x_k, u_k) + w_k,
\end{equation}

where, for any $k\in \mathbb{N}$ :

\begin{itemize}
	\item $x_k \in \textbf{X} \subset \mathbb{R}^d $ is the state space;
	\item $u_k \in \textbf{U} \subset \mathbb{R}^m $ is the control input and $\textbf{U} $ is the control input space;
	\item $w_k$ is the real random variable, which models the unknown part of the system.
\end{itemize}

We model the known parts of a state-space model of the system as $f(x_k,u_k)$ and 
think of the unknown parts of the model as an additive state-dependent disturbance. 

\subsection{Gaussian Process}\label{GP}

Assuming jointly-Gaussian probability distribution of the disturbances and inputs, 
we model entire system as a stochastic process \textit{over the states}. The method employed for disturbance estimation is a Gaussian Process (GP) regression \cite{Rasmussen}:

\begin{equation}
p_{x_{k},u_k}(x_{k+1})  \sim \mathcal{GP} (f(x_k,u_k ) + \mathbb{E} (w_k), k(x_{k+1}, x') ),
\end{equation}

where mean $E(w_k)$ and covariance $k(x_{k+1}, x')$ describe unknown part and noise of the system. They are chosen to capture the characteristics of the model (linearity, periodicity, etc), and defined by a set of hyperparameters $\theta_p $.Inspired by \cite{conf/cdc/AkametaluKFZGT14} we use the squared exponential covariance function to define 
the kernel $k(x_{k+1}, x')$:

\begin{equation}\label{SEcovsigmaf1}
k(x,x') = \sigma_p^2 exp( - \frac{(x - x') L^{-1} (x - x')}{2} ),
\end{equation}

where $L$ is a diagonal matrix, with $L_i$ as the $i$th diagonal
element being the squared exponential’s characteristic
length for the $i$th state, $\sigma_p^2$ being the signal variance, $\sigma_p^2$ being the measurement
noise variance and $x'$ is already given input in the system. Take in account, that covariance kernel function of output of the system is defined trough the inputs. In such a way, it can be shown that the squared exponential covariance 
function corresponds to a Bayesian linear regression model with an infinite number of basis functions, as through the Mercers theorem \cite{konig2013eigenvalue}. All provided coefficients form the hyperparameters $ \theta_p = [ \sigma_n^2, \sigma_p^2, L_1, \ldots, L_n ]$, 
are chosen to maximize the \textit{marginal likelihood} of the training
data set and are thus recomputed for each new batch of data, so-called \textit{MLE-estimate}:

\begin{equation}\label{MLE1}
\hat{\theta}_{MLE} = \underset{\theta}{\arg\max} \sum\limits_{i=1}^n \log p(y_i|x,\theta). \nonumber
\end{equation}

In the Bayesian formalism, we can now specify a prior over the parameters, expressing our beliefs about the prior parameters, just writing the joint prior distribution of the training outputs, $y$ , and the test (predictive) outputs $p_{x_{k},u_k}(x_{k+1})$:

\begin{equation}
\begin{bmatrix}
y \\ p_{x_{k},u_k}(x_{k+1})
\end{bmatrix} \sim \mathcal{N} \left( \begin{bmatrix} 0 \\ f(x_k,u_k )
\end{bmatrix} , 
\begin{bmatrix}
K(X,X) + \sigma_n^2 I & K(X,X_*) \\ K(X_*,X) & K(X_*,X_*) \\
\end{bmatrix}  \right).
\end{equation}

If there are $n$ training points and $n_*$ test points then $K(X,X_*)$ denotes the
$ n \times n_*$ matrix of the covariances evaluated at all pairs of training and test
points, and similarly for the other entries $K(X,X)$, $K(X_*,X_*)$ and $K(X_*,X)$. 
The same way we can obtain joint distribution for one test point $x_*$.

Deriving the conditional distribution, using Gaussian Identities ( see Appendix A in the full version ), we  arrive at the key predictive Equations for Gaussian process regression, and therefore the probability of expected dynamics in \eqref{dynamics1} can be evaluated at any given state $x_*$ for the future reachability analysis:

\begin{gather}
p_{x_{k+1}} \mid X,y,x_* \sim \mathcal{N} ( ( f(x_k,u_k ) +  K(x_*,X) [K(X,X) + \sigma_n^2 I]^{-1} ) y, \\  
K(x_*,x_*) - K(x_*,X)[K(X,X) + \sigma_n^2 I]^{-1} K(X,x_*) ). \nonumber 
\end{gather}\label{inference1}

Regarding safety constraints, technics for reachability analysis 
used in \cite{article} for Invariance Stochastic Problem has been chosen. Due to the marginalization 
of the probability distribution over the safe states we can build a reward function with the dependence 
of safety constraints, in other words, the introduced later algorithm has zero reward outside of the safe region, 
meanwhile maximizing chances to stay in. We, therefore, collect data samples through interactions with the system and estimate a model and disturbance based on GP regression.

\subsection{Reachability Analysis}\label{RA}

We seek for the following class of controls $\mu$:

\begin{equation}
U = \mu : \{ X \times \mathbb{N} \rightarrow \textbf{U} \},
\end{equation}

namely, the class of time–varying feedback functions.
Let $U_N$ be the class of control inputs sequence 
$\pi =\{\mu_k\}_{k=0,1,\ldots,N}$ such that 
$\mu_k \in U$, for any $k = 0, 1, \ldots, N$.
Any $\pi \in U_N$ is \textit{called control policy}.
Given $N \in \mathbb{N}$, and $\pi = \{\mu_0, \mu_1, \dots, \mu_{N-1}\} \in U_{N-1} $, set

\begin{equation}
\pi^k = \{\mu_k, \mu_{k+1}, \dots, \mu_{N-1}\},
\end{equation}

for any $k \in \mathbb{N}$. \\

Given a \textit{safe} set $S \in \textbf{X}$ representing the set of ‘good’ states within which the state evolution of system \eqref{dynamics1} must evolve, our problem is to find a control policy that maximizes the probability of the state 
$x_k$ to be in $S$, for any time $k$ within a finite time horizon $N$. 

More formally, let $(\varOmega,\mathcal{F} , P)$ be the probability space
associated with the system. The Stochastic Invariance Problem we reformulate as follows.

\begin{problem}\label{problem11}
	Given a finite time horizon $N \in \mathbb{N}$ and a safe set
	$S$ subset of $\textbf{X}$, find the optimal control policy
	$\pi^* \in U_{n-1}$ that maximizes
	\begin{equation}\label{probquantity11}
	P(x_k \in S, \forall k = 0,1,\ldots,N)
	\end{equation}
\end{problem}

In the next section, we used Lemmas and Propositions, proven in \cite{article}, changing the framework adapted for our case with the presence of safe set of states $S$ and a target set $X_T$ for implicitly define probability quantity from Problem \ref{problem11} and construct a recursive optimization algorithm, that enables the computation of optimal control policy $\pi^*$. (See Appendix in section \ref{AppendixA})

\section{Proposed Approach}

In accordance with Problem \ref{problem11}, reformulate optimal control problem for stochastic systems, as our predictive dynamic described by \eqref{dynamics1}, with integrated safety approach:

\begin{problem}\label{problem3}
	Given a non-linear stochastic discrete-time dynamical control system $x_{k+1} = f(x_k,u_k,w_k)$, 
	initial state $ x_0 \in \textbf{X} $, the target set of states $X_T \in \textbf{X}$, 
	set of safe states $S \subset \textbf{X} $ and finite time horizon $N \in \mathbb{N} $,  
	find the optimal control policy $\pi^* \in U_{n-1}$ that maximizes
	
	\begin{equation}\label{probquantity2}
	P(x_k \in S, \forall k = 0,1,\ldots,N-1; x_N \in X_T \mid x_0 ).
	\end{equation}
\end{problem}

We now give the main result of the paper \cite{article}: an algorithm that enables the computation of optimal control policy $\pi^*$ 
solving the Reachability Problem \eqref{problem3} just by greedy solving recursively optimization problem for every step in finite time horizon $N$. Additionally, we reformulate it for the probability distribution, given by GP regression estimation in \eqref{inference1}.

\begin{theorem}\label{mainthm1}
	The optimal value of the problem \ref{problem3} is equal to
	
	\begin{equation}
	p^*(N) = \int_S J_0 (x) dx, \nonumber
	\end{equation}
	where $J_0(x)$ is given by the last step of the following recurrence 
	algorithm,
	
	\begin{gather}\label{recurcive1}
	\begin{dcases}
	J_N (x) = I_{X_T} (x), \ \ \ k = N  \\ 
	J_k (x) = \sup_{u_k \in \textbf{U}} \int_S J_{k+1} (z) p_{f(x,u_k)} (z) dz, \\ k = N-1, N-2, \ldots, 0.
	\end{dcases}
	\end{gather}
	
	Furthermore, if $\hat{\mu}_k(x) = \hat{u}_k $ maximizes the right-hand side
	of Equation \eqref{recurcive1} for each $ x \in S $ and $k = 0,1 \dots N-1$, then the
	class of policies $ \hat{\pi} = \{\hat{\mu}_0, \ldots, \hat{\mu}_{N-1}\} $ is optimal. 
\end{theorem}

\textit{Proof:} Omitted. See Appendix (section \ref{AppendixB}) with following theory needed. 

\subsection{Exploration}\label{Explorationa1}

Inspired by \cite{Lee2010_2010arXiv1004.4027G}, \cite{BayUncConst2014arXiv1403.5607G}, it has been chosen methodology for exploration purposes, which is useful in deriving more precise model through simulations with randomness. After a predefined number of steps, 
we relocate the target set $X_T$ inside the safe set of states $S$ to be close to the state, taking in account marginalization over all the control inputs, with the maximal predicted variance, or the maximal difference between upper and lower confidence bounds in other words. Define $\hat{S} $ as $\hat{S} =  S - r $ for a fixed radius $r$. Let $U_f$ be a set of feasible control inputs and $x_*$ is the input vector, formed by the combination of states $x$ and control inputs $u$, as it was given before $x_* = \begin{bmatrix} x \\ u \end{bmatrix}$. Define constrained optimization problem: 

\begin{equation}
x_t = \argmax_{x \in \hat{S}} \int_{U_f} K(x_*,x_*) - K(x_*,X)[K(X,X) + \sigma_n^2 I]^{-1} K(X,x_*)  du 
\end{equation}

As a result, we a form new target set $X_T$ as a ball of radius $r$ centered in $x_t$, that allows the safe exploration of the system and proved numerically in the next chapter.

\section{Implementation}

\subsection{Inverted Pendulum}

As a benchmark example consider a damped inverted pendulum system with mass $m$, length $l$, and friction
coefficient $b$. The states of the system are the pendulum angle $x_1$ and angular
velocity $x_2$. The system is disturbed by an additive state-dependent disturbance
$d(x)$. The description of dynamics given by \cite{doya2000reinforcement}:

\begin{gather}\label{contSystem1}
\begin{cases}
\dot{x_1} = x_2 + d_1(x) \\
\dot{x_2} = \frac{1}{ml^2} u + \frac{g}{l}\sin(x_1) - \frac{b}{m} x_2 + d_2(x)
\end{cases}
\end{gather}

All constants are assumed to be positive.

We would like to approximate this system with the discrete time system in state space 
so that $x_k := x(kT) \approx \hat{x_k}$, where $T$ is a sampling time. 
Notice that $\hat{x_k}$ might not be equal to $x_k$ because there are 
infinite input signals $u(t)$ which have the values at the sampling points $kT$,
therefore, in general, they give rise to different output signals $x(t)$. As so, it is impossible
to find an equivalent discrete-time system such that $\hat{x_k} = x(kT)$ for \textit{any} input signal $u(t)$. 
Therefore, the best we can hope for is to find an approximation of the time derivative of $\dot{x}(t)$.
We can approximate $ \dot{x}(t) = \lim_{\varepsilon \rightarrow 0} \frac{x(t + \varepsilon) - x(t)}{\varepsilon}$ as:

\begin{equation}
\dot{x}(t) \approx \frac{x(t+T) - x(t)}{T} \nonumber
\end{equation}

for small enough values of $T$. The approximation associated with these substitutions is known as Forward Euler Approximation
for state space models:

\begin{gather}\label{modell}
\begin{cases}
\frac{x_{k+1}^1 - x_k^1}{T} = x^2_k + d_1(x^1_k) \\
\frac{x_{k+1}^2 - x_k^2}{T} = \frac{1}{ml^2} u_k + \frac{g}{l}\sin(x_k^1) - \frac{b}{m} x_k^2 + d_2(x_k^2)
\end{cases}\Leftrightarrow
\begin{cases}
x_{k+1}^1 = T x^2_k +  x_k^1 + T d_1(x^1_k) \\
x_{k+1}^2 = x_k^2 +\frac{T}{ml^2} u_k + \frac{g T}{l}\sin(x_k^1) - \frac{bT}{m} x_k^2 + T d_2(x_k^2)
\end{cases}\nonumber
\end{gather}

It is hereby important that the time step $T$ is small enough to actually allow to the learning 
agent to react on the changes, otherwise could be the position of "synchronization" 
between state change and the reaction of the system. From the other side, 
a small time step does not allow to computationally proceed the algorithm, 
so this question under further consideration.

Since in the problem solve we are using programming tools, defining
the set of states $S$ and the set of actions $U$ implies discretization over
the intervals $ [ x_{\min} ; x_{\max} ] $ 
and $ [ u_{\min}; u_{\max} ] $, 
where $ x_1 $ is a circular state and should be in the interval is not larger than its period $ 2 \pi $. The number
of discretization steps impacts the convergence speed of the learning
algorithm, due to correspondence to the number of evaluation of GP function during integration, 
and was chosen around $nn = 40$ for each state dimension so that we end
up with $1600$ evaluations in total for one step.

During the implementation of this technics has been used Matlab software with additional GPML toolbox for model Gaussian Process in a manner described \cite{Rasmussen}. Note, that any further call to the GP function corresponds to one of the aspects, as inference or likelihood estimation through this toolbox\cite{rasmussen2010gaussian}, \cite{rasmussen2016gpml}.
Pseudo-code for this problem solution is introduced in Algorithm \ref{alg:algorithm1}, where apart from the data given above, 
it is required to define $N$ as the number of steps for time horizon considered, $n$ as the total number of steps to be done by system, 
$hyp$ as initial conservative hyperparameters, as variance and constants for polynomials, and define the function of the model, which is actually corresponds to \eqref{modell} with additional bounds on total state space $S$ already incorporated. 

\begin{algorithm}[H]
	\begin{algorithmic}
		\Require  \textbf{integer} $N$, $n$, $nn$; \textbf{real array}  $hyp$; \textbf{external function} $y$; \textbf{real} $T$, $mass$, $l$,  $g$, $b$, $ u_{\max}$, $u_{\min}$, $x_{\max}$, $x_{\min}$, $X_{t}$ 
		\State S(1,:)  $\leftarrow$ linearly spaced vector from $x_{\min}(1)$ to $x_{\max}(1)$ with $nn$ points
		\State S(2,:)  $\leftarrow$ linearly spaced vector from $x_{\min}(2)$ to $x_{\max}(2)$ with $nn$ points  
		\State $X \leftarrow [0,0.1,0]$ \Comment{ $X \neq 0$ for the further inverse }
		\For{i=1}{ N }
		\State $Y(i,:) \leftarrow y(X(i,:)) $
		\State Update  $X_t$ 
		\If{ $mod(i,10) = 0$ } 
		\State $hyp \leftarrow \min$ (GP negative likelihood($X$,$Y$, $hyp$))
		\EndIf
		\State $k \leftarrow 0$ \Comment {Prediction steps count} 
		\State $x_0 \leftarrow Y(i,:)$ \Comment {actual state}
		\State Parameters $\leftarrow$ ($x_0$, $hyp$, $X$, $Y$, $N$, $S$, $X_{t}$, $u_{\max}$, $u_{\min}$, $k$)
		\State $[J, u_{opt}] \leftarrow$ Pattern Search (Cost( Parameters ), $u_{\max}$, $u_{\min}$ )
		\State $X \leftarrow [X \ ; \ Y(i,:) \ \ u_{opt} \ ]$
		\EndFor
	\end{algorithmic}
	\caption{Learning}
	\label{alg:algorithm1}
\end{algorithm}

The function Cost, used above, is described by Algorithm \ref{alg:algorithm2}.

\begin{algorithm}[H]
	\begin{algorithmic}
		\Require  \textbf{integer} $N$, $k$; \textbf{real array} $hyp$, $Y$, $X$, $S$; \textbf{real} $x_0$, $ u_{\max}$, $u_{\min}$, $X_{t}$ 
		\If{$k = N-1 $}
		\For{ i=1 }{ size ( $X_t$, 1 ) } 
		\State $p_f \leftarrow$ GP predict ($X$, $Y$, $x_0$, $X_t$ )
		\State $J \ \leftarrow \ J \ + \ (X_t(i+1) \ - \ X_t(i)) \times p_f$
		\EndFor	
		\Else 
		\For{i=1}{ length ( $S$, 1 ) } 
		\State $p_f \leftarrow$ GP predict ($X$,$Y$,$x_0$,$S(i,:)$ )
		\State $k \leftarrow k+1$
		\State Parameters $\leftarrow$ ($x_0$, $hyp$, $X$, $Y$, $N$, $S$, $x_{t\max}$, $x_{t\min}$, $u_{\max}$, $u_{\min}$, $k$)
		\State $J_{k+1}  \leftarrow$ Pattern Search (Cost( Parameters ), $u_{\max}$, $u_{\min}$ )
		\State $J \leftarrow J +  (S(i+1) \ - \ S(i)) \times J_{k+1} \times p_f$
		\EndFor 
		\EndIf
	\end{algorithmic}
	\caption{ Cost function }
	\label{alg:algorithm2}
\end{algorithm}

\subsection{Results}

\begin{figure}[H]
	\centering
	\begin{subfigure}[b]{0.48\textwidth}
		\includegraphics[width=1\linewidth]{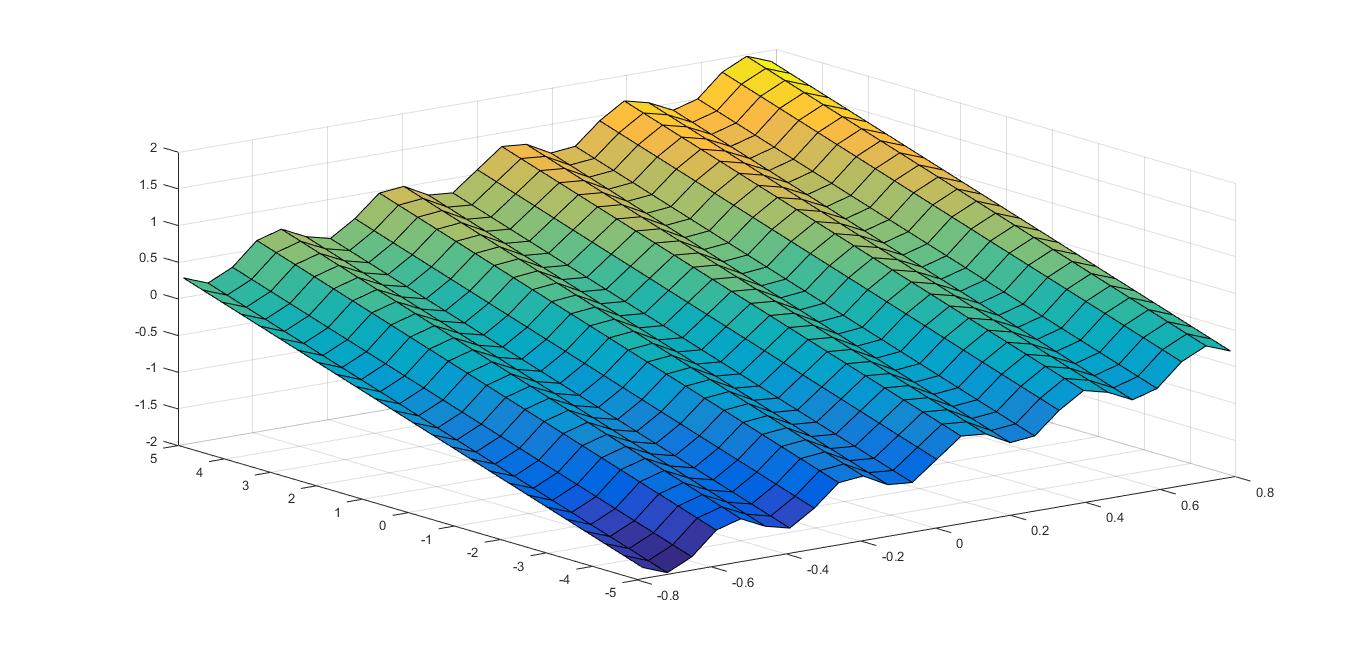}
		\caption{True output $x_1, \ rad$ }
		\label{fig:originalx1}
	\end{subfigure}
	\begin{subfigure}[b]{0.48\textwidth}
		\includegraphics[width=1\linewidth]{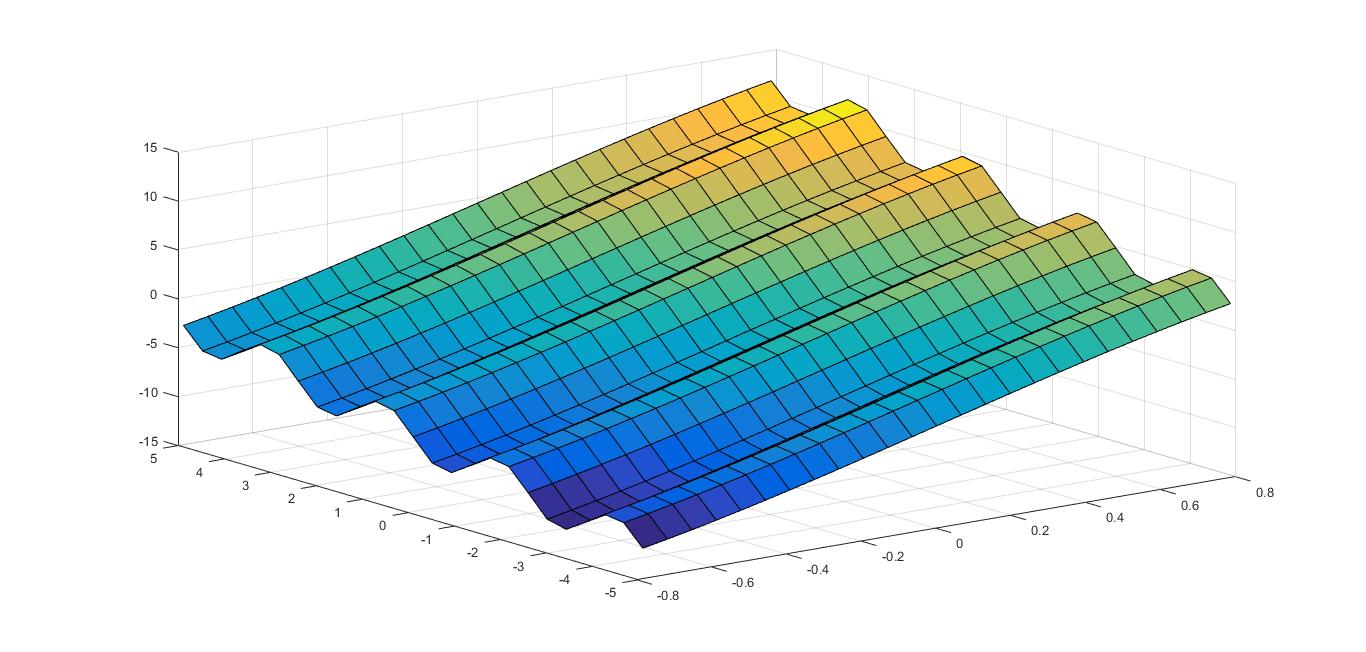}
		\caption{True output $x_2, \ \frac{rad}{s}$ }
		\label{fig:originalx2}
	\end{subfigure}
	\caption{Real output of the system in dependence on input $x_1 \in [-\frac{\pi}{2}\ rad; \ \frac{\pi}{2} \ rad]$, $x_2 \in [-5 \ \frac{rad}{s}; \ 5 \ \frac{rad}{s}]$ and $u = 0 \ N$}
	\label{OriginPic}
\end{figure}

The inverted pendulum system has been modeled as described above with physically meaningful state set bounds  
$x_1 \in [-\frac{\pi}{2}\ rad; \ \frac{\pi}{2} \ rad ]$ and $x_2 \in [-10 \ \frac{rad}{s}; \ 10 \ \frac{rad}{s} ] $. 
The state-space has been discretized with $40$ steps in each
dimension and the action space bounds have been set $ u \in [-5 \ N; \ 5 \ N]$ as feasible. 
In addition, we form the safe set $S$ as restriction over $ x_1 \in [\frac{\pi}{4} \ rad; \ \frac{\pi}{4} \ rad]$, which 
corresponds to $90 \ \deg $, and the initial target set $X_T$ as a ball with radius $r = \frac{2\pi}{100}$ : $ x_1 \in [-\frac{2\pi}{100} \ rad ; \  \frac{2\pi}{100} \ rad] $, equal 
to around $ 2,3 \ \deg $. 
The sampling time for the learning loop is chosen as $ T = 0,2 \ s$. Initially, true disturbance introduced
to the system is chosen to be $d_1(x_1) = \frac{1}{5} \sin (20 x_1)$ and $d_2 (x_2) = 2 \sin (3 x_2)$ and graphically presented in Figure \ref{OriginPic}. This picture presents the real output of the system with the dependence of combination of the input state $x_1 \ rad$ and $x_2 \  \frac{rad}{s}$, given the control input $u = 0$ for reference. Hence, for having the possibility to describe such complex functions, after some empirical considerations, 5-th degree Polynomial mean function has been chosen. As it was described above,  
covariance function is defined in accordance with Eq.\eqref{SEcovsigmaf1}. The set of hyperparameters, in this case, is formed by $\theta_p = [ \sigma_n^2, \sigma_p^2, L_1, \ldots, L_n, h_{1,1}, \ldots, h_{D,1}, \ldots,  h_{D,5} ]$, where $\sigma_p^2$ being the signal variance, $\sigma_n^2$ being the measurement
noise variance, and $L_i$ being the squared exponential’s characteristic
length for the $i$th state and $h_{i,j}$ are coefficients before polynomials. The hyperparameters are
chosen as a solution of maximization of the marginal likelihood of the training
data set and are thus sequentially recomputed for each new batch of data every 8 steps.

\begin{figure}[H]
	\centering
	\begin{subfigure}[b]{0.8\textwidth}
		\includegraphics[width=1\linewidth]{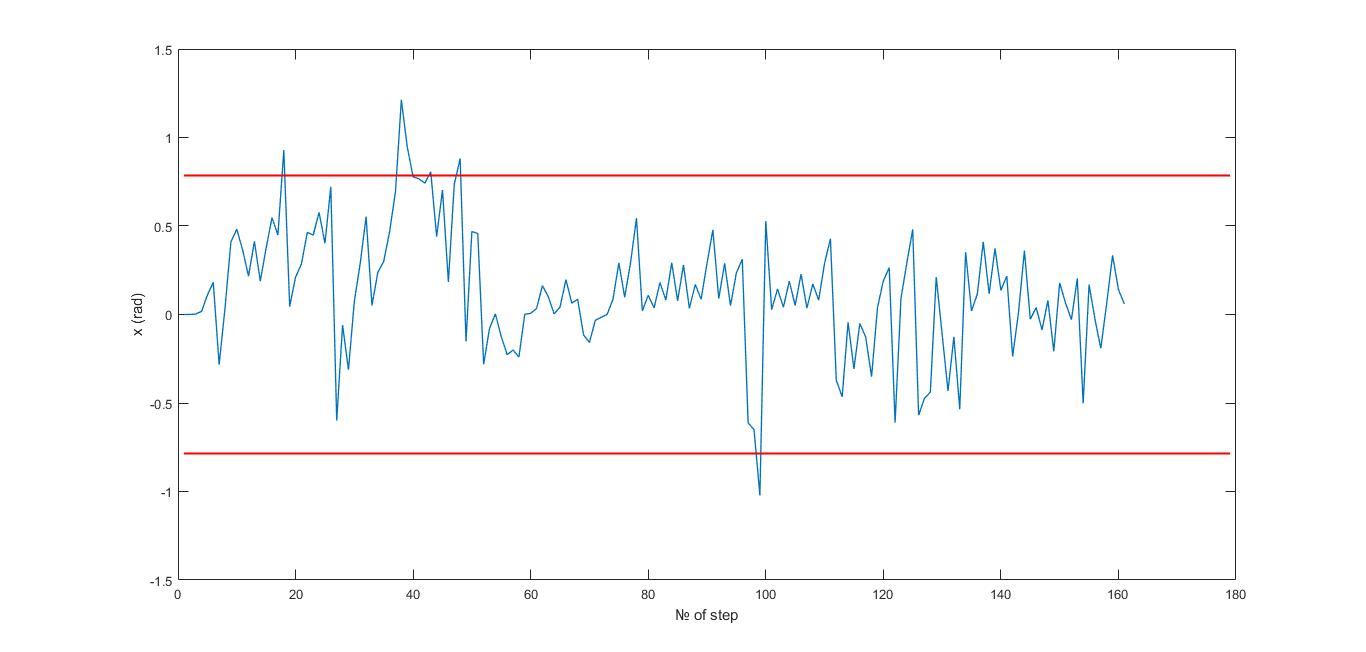}
	\end{subfigure}
	\begin{subfigure}[b]{0.48\textwidth}
		\includegraphics[width=1\linewidth]{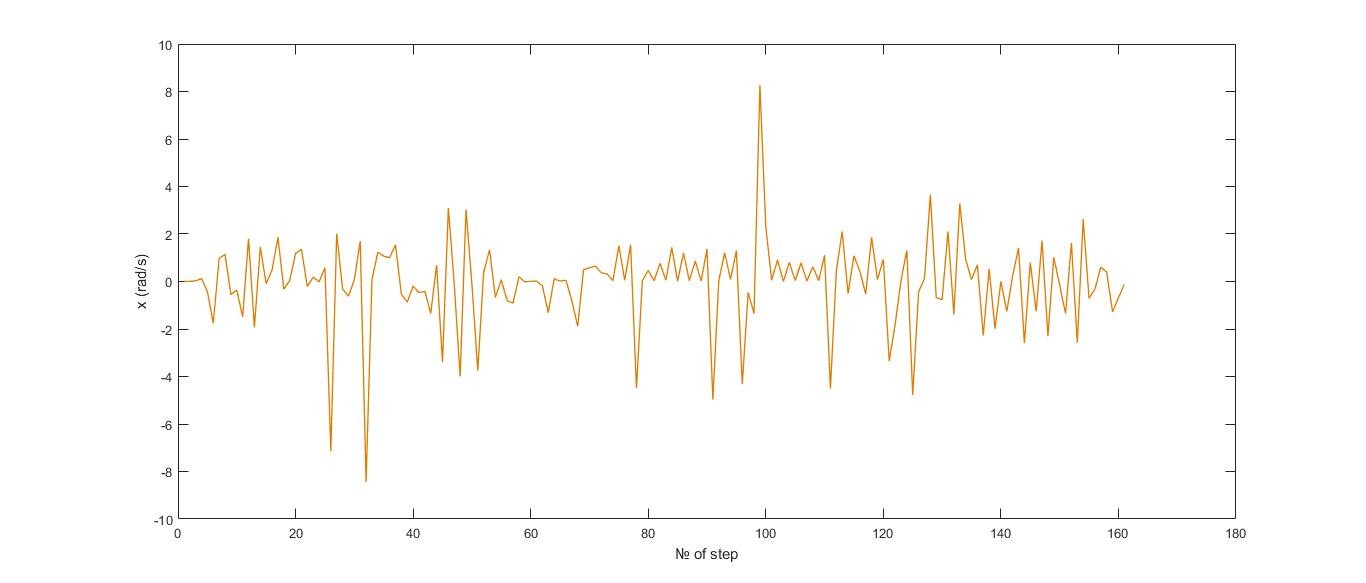}
	\end{subfigure}
	\begin{subfigure}[b]{0.48\textwidth}
		\includegraphics[width=1\linewidth]{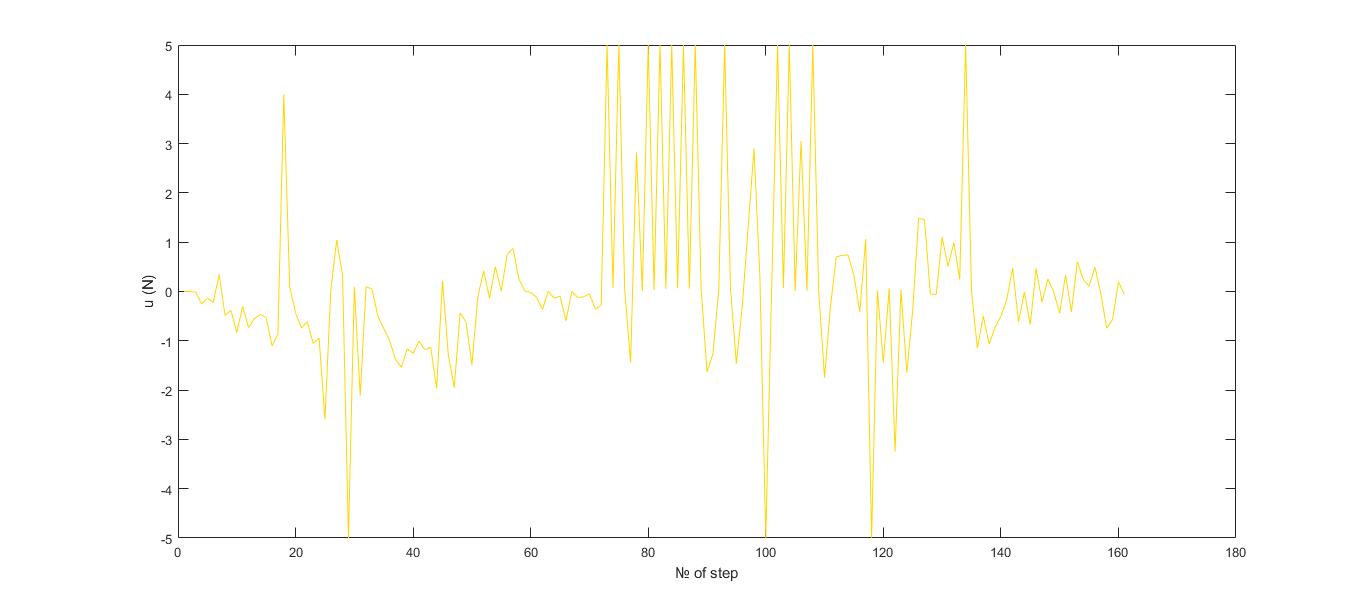}
	\end{subfigure}
	\caption{Resulting dynamic of the system in 4 learning iterations with $40$ steps each. There are presented on top changing of states $x_1 \in [-\frac{\pi}{2}\ rad; \ \frac{\pi}{2} \ rad]$, which is the angle of pendulum in $rad$, and down there are angular velocity $x_2, \ \frac{rad}{s}$ and control input $u, \ N$ accordingly with the number of steps. Additionally, with the red line it is showed safe region $S$}
	\label{DynR0Pic}
\end{figure}

\begin{figure}[H]
	\centering
	\begin{subfigure}[b]{0.48\textwidth}
		\includegraphics[width=1\linewidth]{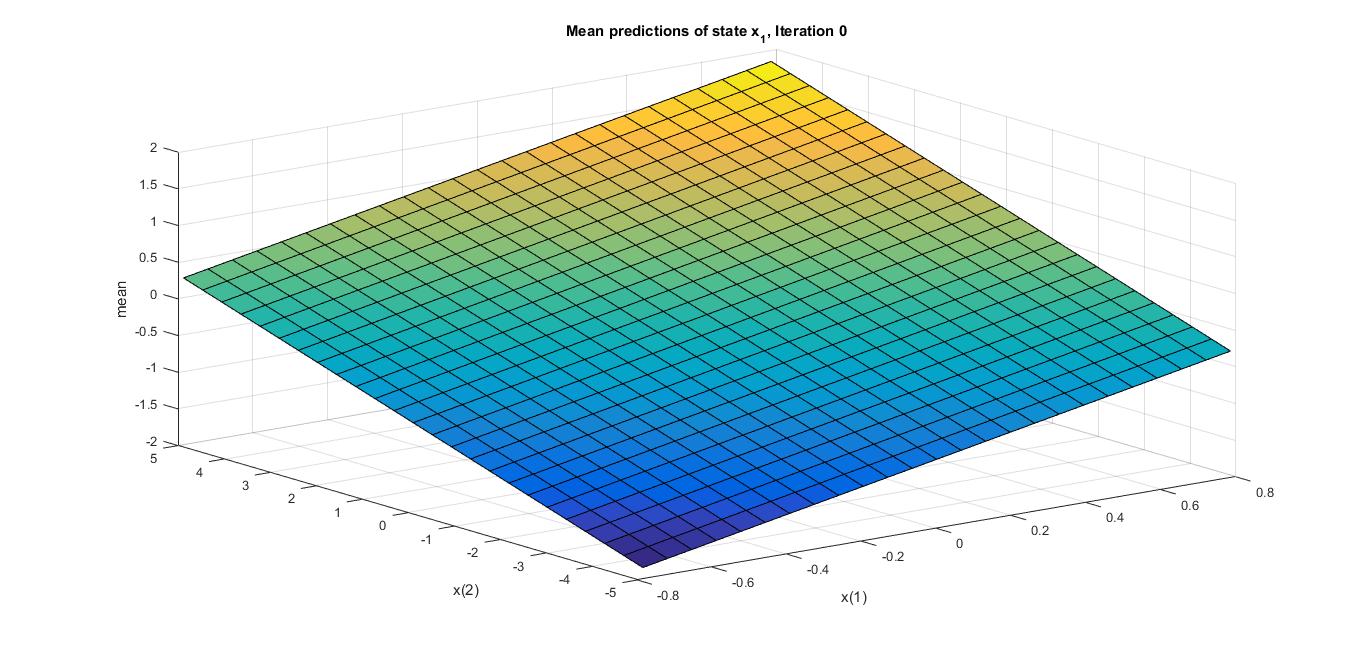}
	\end{subfigure}
	\begin{subfigure}[b]{0.48\textwidth}
		\includegraphics[width=1\linewidth]{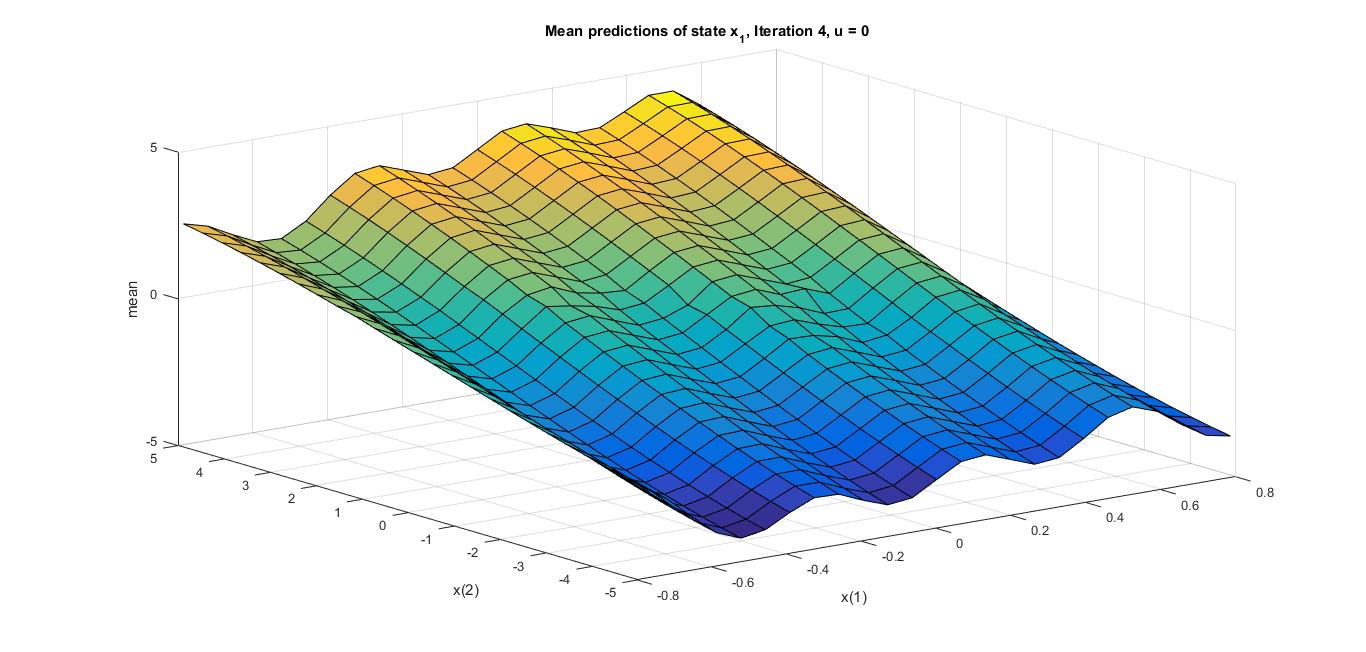}
	\end{subfigure}
	\begin{subfigure}[b]{0.48\textwidth}
		\includegraphics[width=1\linewidth]{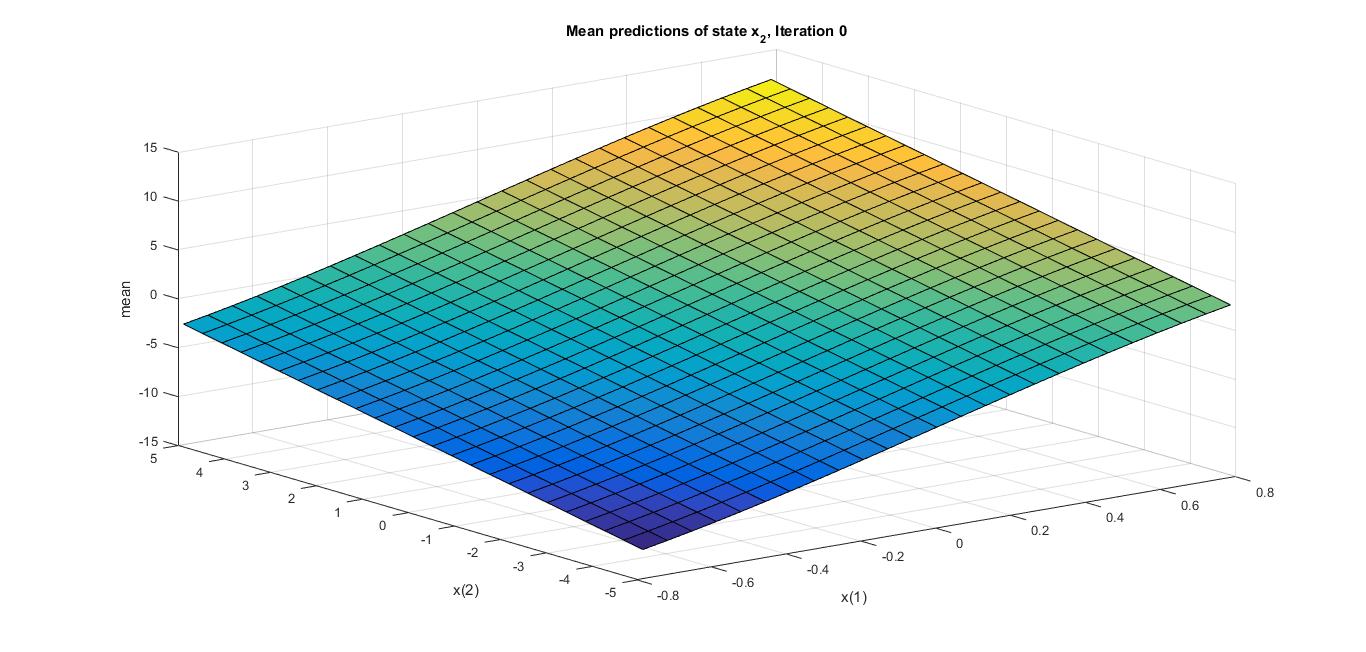}
	\end{subfigure}
	\begin{subfigure}[b]{0.48\textwidth}
		\includegraphics[width=1\linewidth]{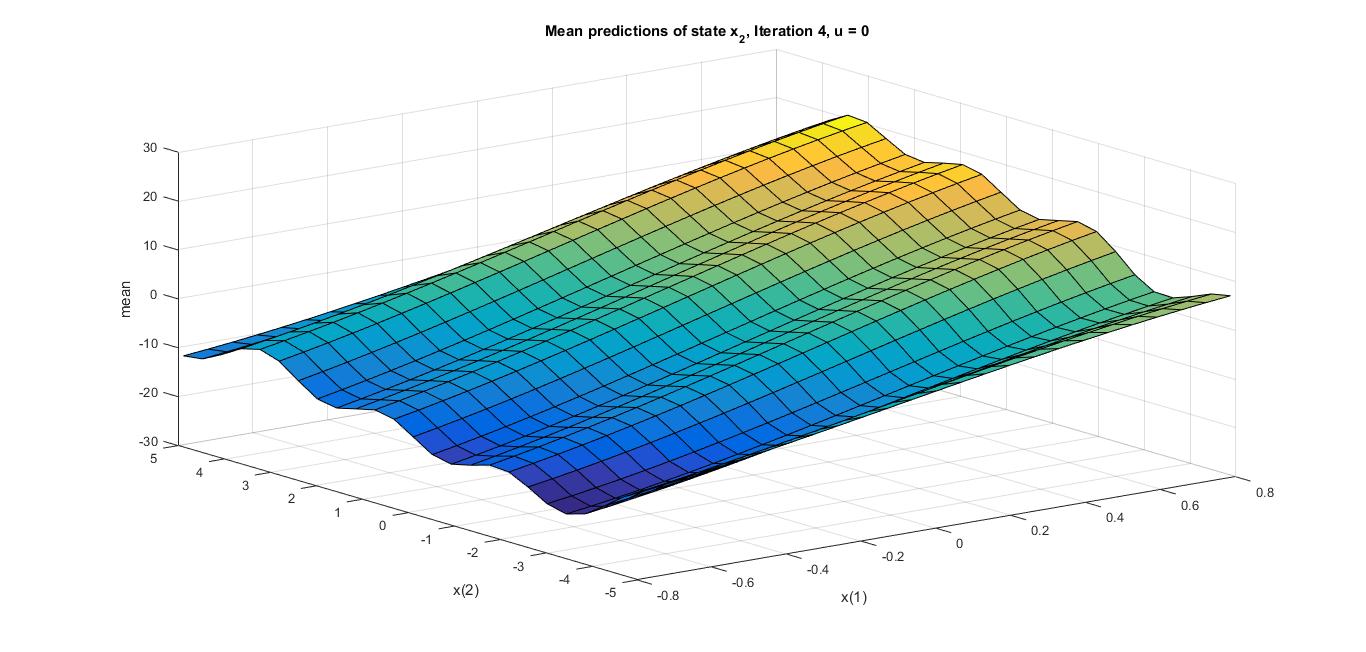}
	\end{subfigure}
	\caption{Resulting mean prediction changes of states $x_1$ and $x_2$ accordingly before (left) and after (right) 4 learning iterations with $40$ steps each.}
	\label{MeanR0Pic}
\end{figure}

\begin{figure}[H]
	\centering
	\begin{subfigure}[b]{0.48\textwidth}
		\includegraphics[width=1\linewidth]{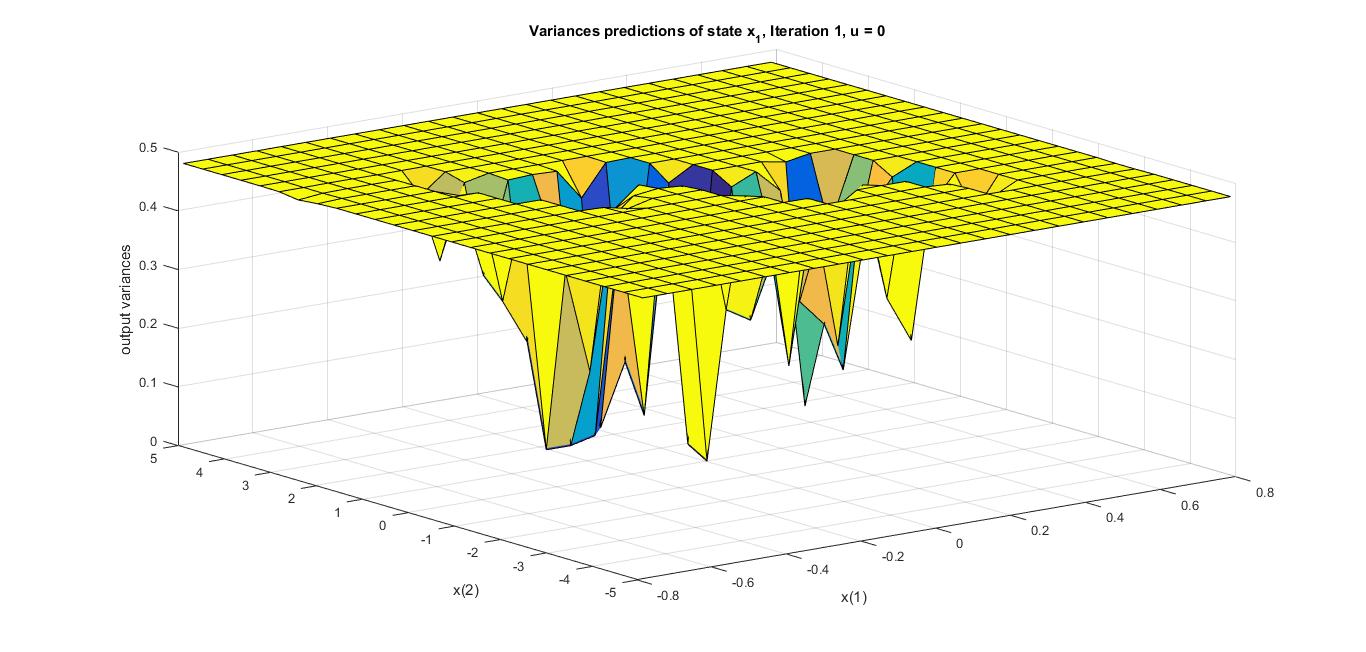}
	\end{subfigure}
	\begin{subfigure}[b]{0.48\textwidth}
		\includegraphics[width=1\linewidth]{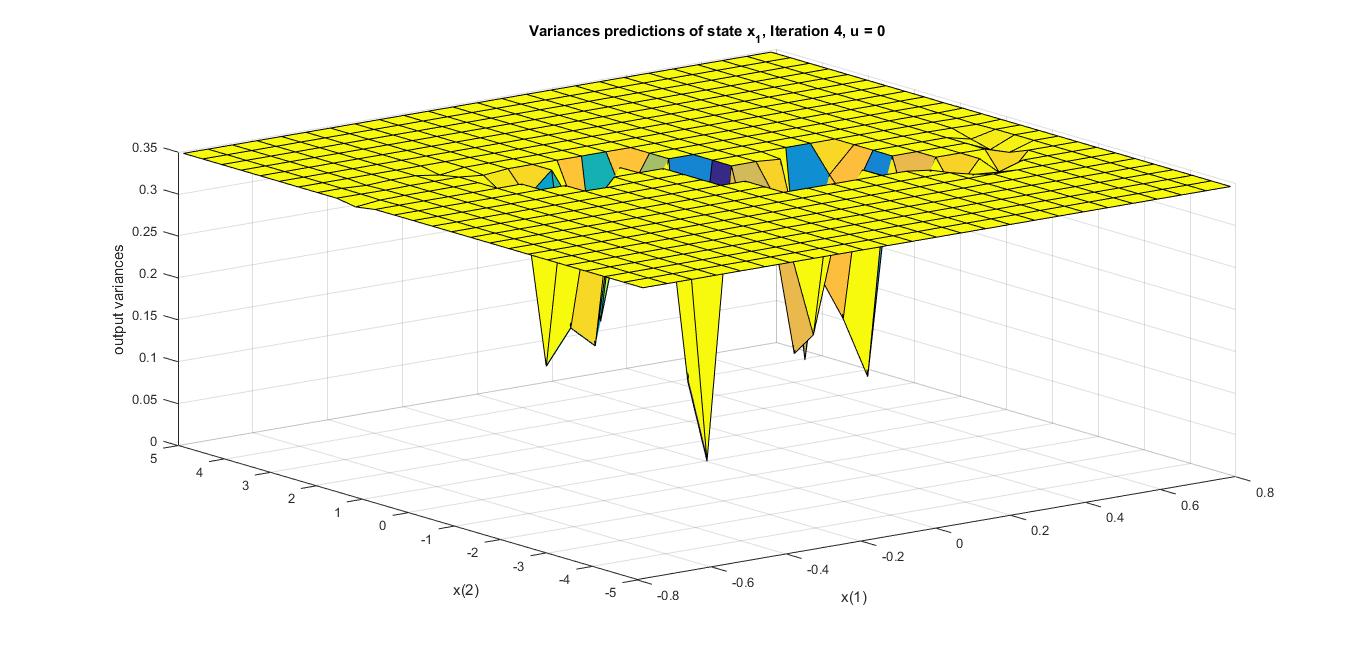}
	\end{subfigure}
	\begin{subfigure}[b]{0.48\textwidth}
		\includegraphics[width=1\linewidth]{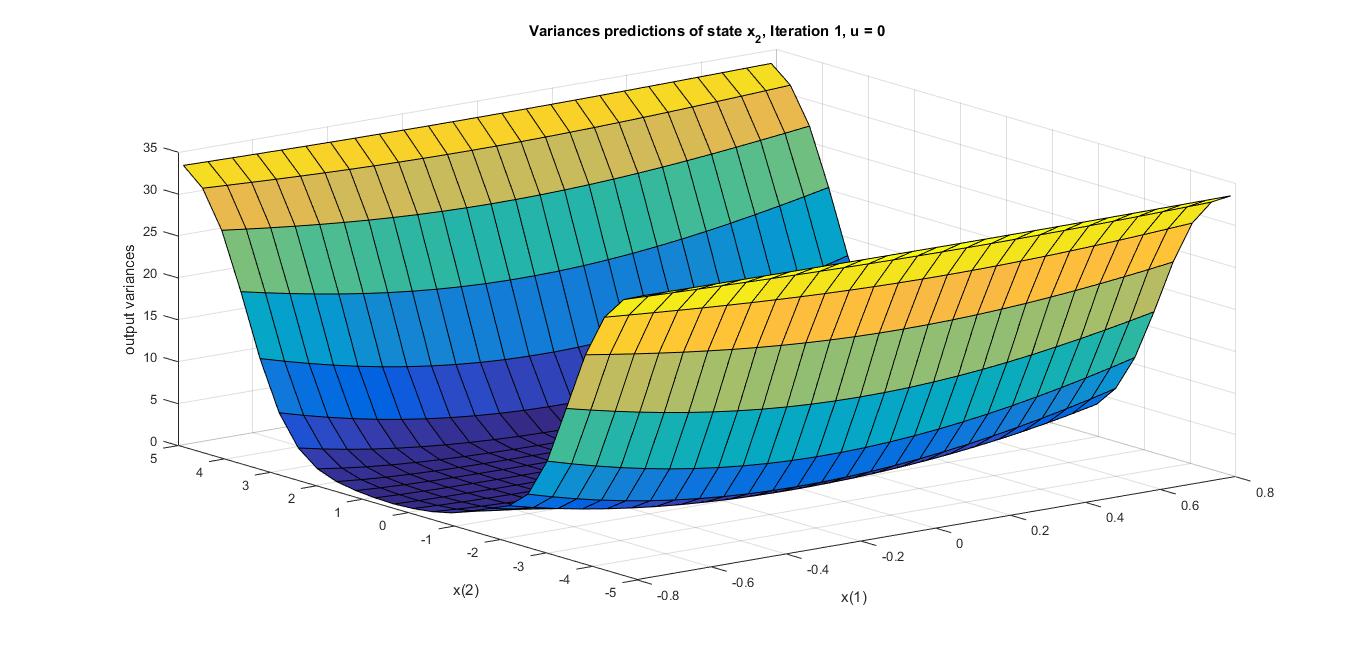}
	\end{subfigure}
	\begin{subfigure}[b]{0.48\textwidth}
		\includegraphics[width=1\linewidth]{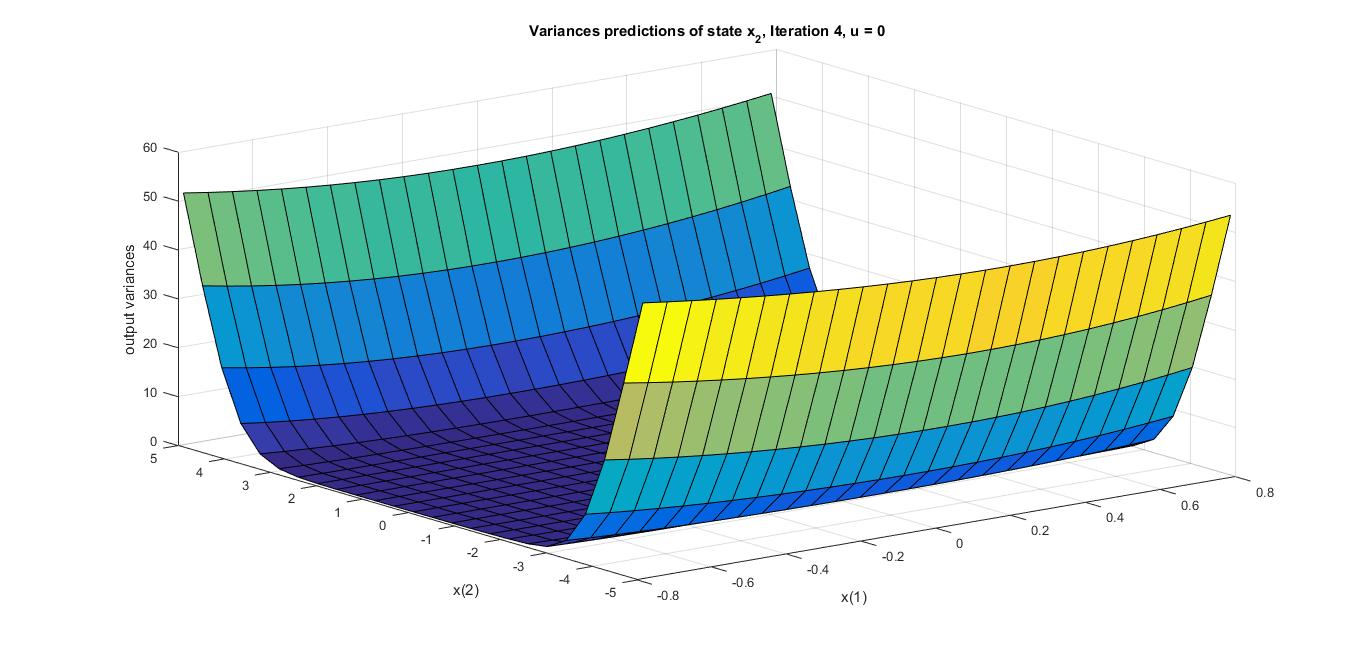}
	\end{subfigure}
	\caption{Resulting variances prediction changes in the logarithmic scale of states $x_1$ and $x_2$ accordingly after first (left) and after 4 (right) learning iterations with $40$ steps each}
	\label{VarR0Pic}
\end{figure}

In what follows, the results from four learning iterations with every $40$ steps are presented. In this setting, exploration mechanics has not been incorporated. After each iteration, a new disturbance estimation, hyperparameters, and number of failures are carried out. The dynamics of the system is given in Figure \ref{DynR0Pic}, where in addition with the red line showed boundaries of safe set $S$, so it is possible to define whether system attempts to fail. Evolution of expectation and variances of GP over the state-space are located on the Figures \ref{MeanR0Pic} and \ref{VarR0Pic}.

\begin{figure}[H]
	\centering
	\begin{subfigure}[b]{0.49\textwidth}
		\includegraphics[width=1\linewidth]{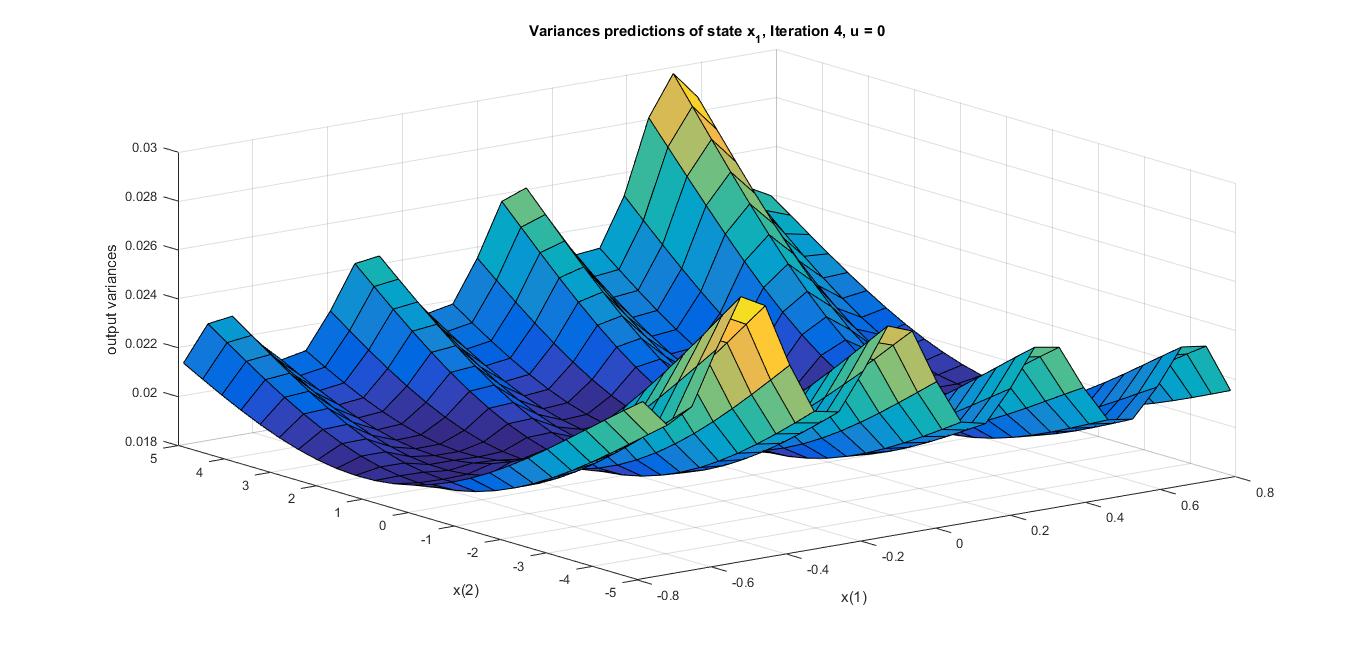}
		\label{fig:24r1}
	\end{subfigure}
	\begin{subfigure}[b]{0.49\textwidth}
		\includegraphics[width=1\linewidth]{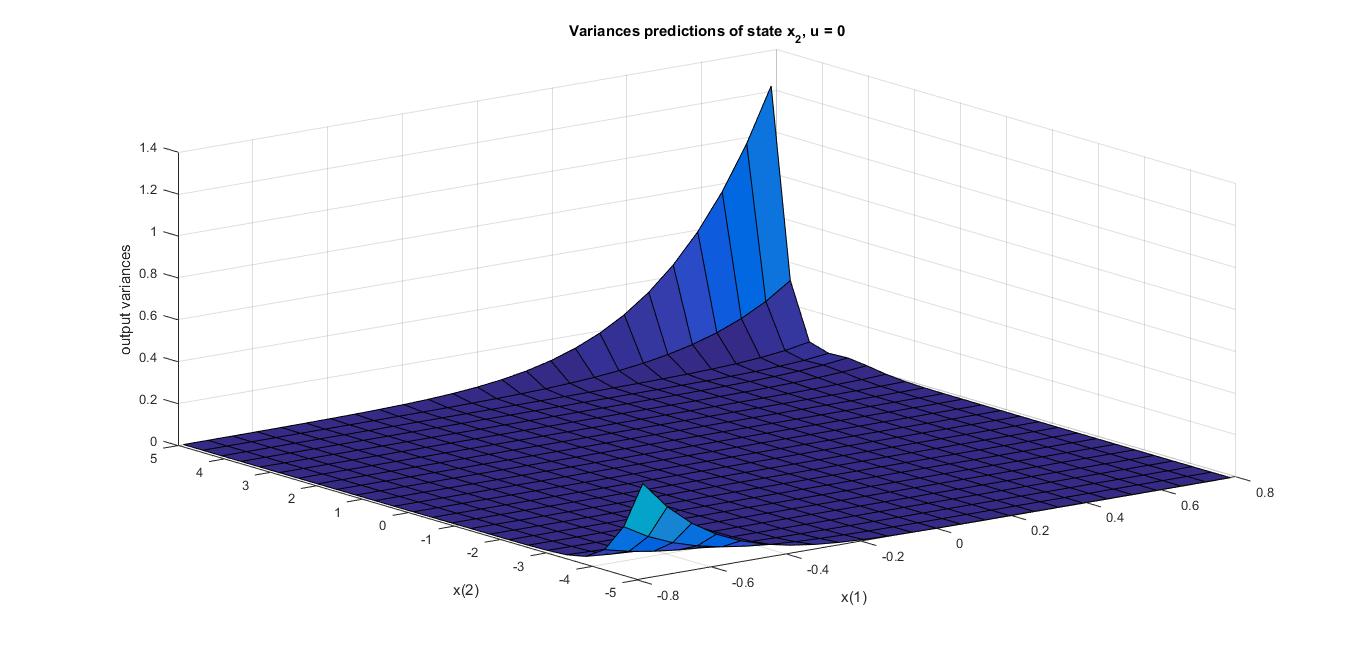}
		\label{fig:RR4}
	\end{subfigure}
	\caption{Resulting variances prediction in logarithmic scale for states $x_1$ and $x_2$ accordingly after 4 learning iterations with $40$ steps each}
	\label{VarR1Pic}
\end{figure}

It is clear, that algorithm managed to learn disturbance and stay in the safe region. But, observing the variances, we can say that not all the system is identified and further exploration to ensuring safety needed. For this purpose has been introduced exploration set up. Significant numerical changes in mean function values were not noticed, but we see great improvements for the variance after 4 iterations, they are given in Figure \ref{VarR1Pic}. The same way safety constraints were satisfied after the second iteration done, but as it can be seen, more precise knowledge of the system is obtained, since even the order of variance become up to $10^{-2}$, which corresponds to significantly decreased uncertainty.

\begin{figure}[H]
	\centering
	\begin{subfigure}[b]{0.72\textwidth}
		\includegraphics[width=1\linewidth]{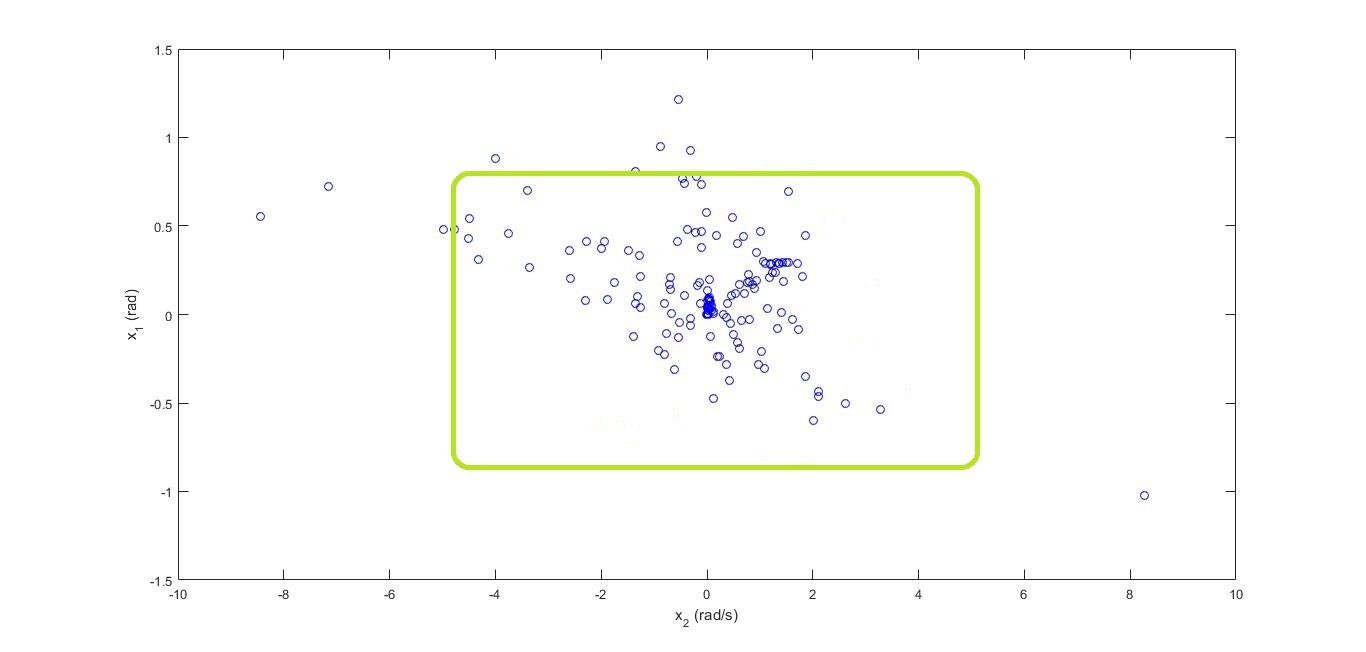}
		\caption{Realization without exploration.}
	\end{subfigure}
	\begin{subfigure}[b]{0.72\textwidth}
		\includegraphics[width=1\linewidth]{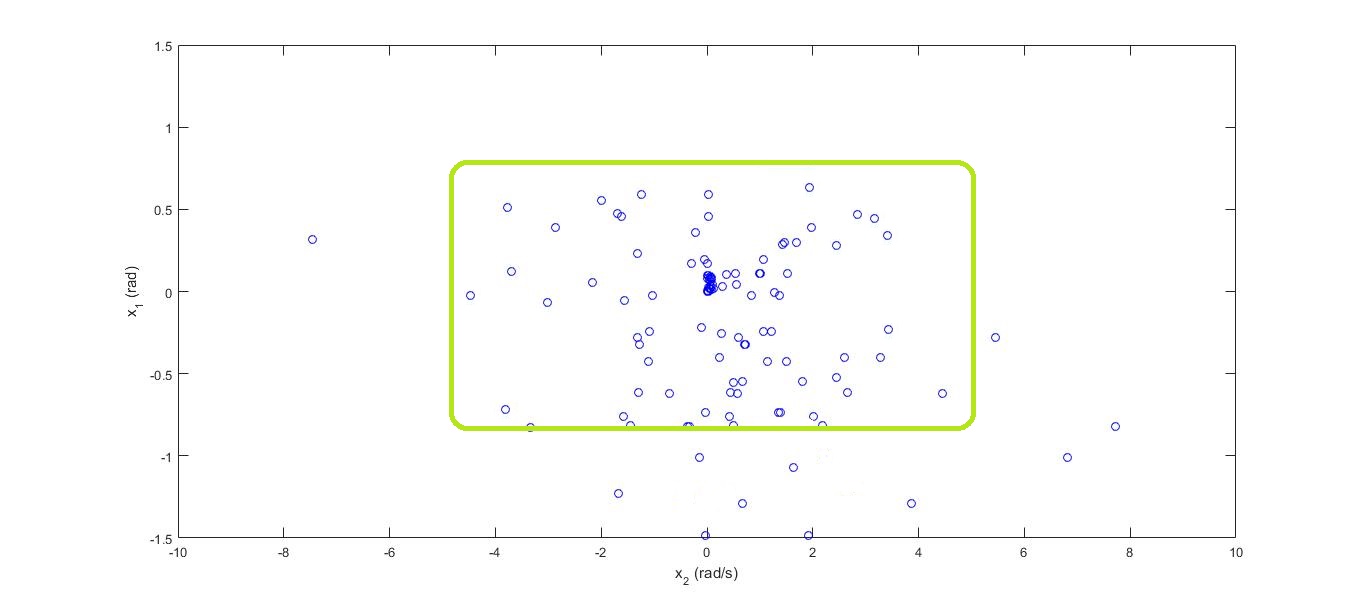}
		\caption{Realization with exploration.}
	\end{subfigure}
	\caption{ Realization of samples with and without exploration. In the lower figure, the spread of the samples is clearly better, whereas in the upper figure most samples are concentrated around the origin and one line.}
	\label{CompR1R0Pic}
\end{figure}

Additionally, here is presented comparative statistics of the system with incorporated exploration and without one. On the Figure \ref{CompR1R0Pic} are marked states, visited during the learning process in both settings. With the green rectangle it is shown the boundaries of the safe set $S$, in such a way we can quantitatively display accuracy of the learning algorithm. It is possible to note, that even if incorporation of exploration gives a control strategy more prone to failures, but this in future gives us less variance and, in fact, more confident knowledge about the system. 

\section{Conclusions and future work}

Applying Bayesian learning methods to control tasks is a promising approach
to overcome the strong dependence of traditional model-based control methods on
accurate models especially for non-linear systems. Model-free learning
techniques have been extensively studied and proven their value
in various applications. However, the problem of how to satisfy constraints during
the learning process has not yet been addressed. 
Therefore, Reachability Analysis has been incorporated into the algorithm to ensure safety
during the learning process as well as target achievement. By modeling the unknown parts of the state-space
model as an unknown additive disturbance, this
method provides a way to work with a stochastic system for further optimizing the control policy. 
Combining Bayesian learning and exploration analysis provides a way 
to regard safety learning to control a system with uncertain dynamics. 
Compared to the approach presented in \cite{conf/cdc/AkametaluKFZGT14}, the method has been used 
totally different reachability analysis technic \cite{article} to safely define control policy and 
has been extended to incorporate exploration. We compared the approaches with
exploration and without exploration, and find that both, policy learning and disturbance
estimation, can be considerably improved by encouraging exploration.

However, the proposed algorithm needs improvements. In this paper the inverted pendulum system has been studied, which has the advantage
of being easy to analyze and illustrate. But even for this system, in spite of all the work on optimization, real processing time does not let us apply it in real-world system. For reference, with $N =1$ time horizon, evaluation of one step on $1.7 Ghz$ frequency processor takes from $1$ to $10$ seconds, depending on the size of the bath, and in the working regime, with $N = 2$ it takes up to $200$ seconds, which shows us possibilities to apply it only in not real-time systems. These issues are due to the necessity of taking the invert of the matrices, size of which is dependent of the size of labeled data provided and also due to the usage of "off-the-shelf" optimizers able to work with stochastic functions. Thereby, for the future work it is proposed to work with finite state discretization \cite{doya2000reinforcement}, sparse matrices \cite{Rasmussen} and incorporation in advance learning algorithms, as Reinforcement learning \cite{conf/cdc/AkametaluKFZGT14}. But, speaking about the scalability of the model, convergence speed would be decreased considerably when the number of states increases \cite{conf/cdc/AkametaluKFZGT14}. 
Secondly, the disturbance estimation could possibly be improved. The GP regression is implemented with batches
of samples increasing with every step is done. The disturbance estimation might be improved considerably by employing a recursive
method that takes all recorded samples into account the same way learning process is proceeding. It is not trivial
how to implement GP regression in a recursive manner because the method relies
on re-computing the covariances for each new input.
Finally, the most significant improvement could be made
by performing formal safety guarantees for the whole algorithm. This is, however, a very
difficult matter because we come to relatively safe development of the process after some boundary of knowledge of the system, which is not trivially defined.

In summary, remark that ideas from this paper could serve as a good starting point for future research. As pendulum, however, not really is the "safety-critical applications", later we consider to approach to pump-schedule problem as a strictly constrained problem, which is close to real-world task. Even though some adjustments should be made, i.e. engagement modern learning technologies, the approach of safe learning for control applications is really interesting and could overcome some of the limitations of traditional control theory. 

\section{Appendix}\label{Appendix}

\subsection{Preliminary results}\label{AppendixA}

The first result makes explicit in the probability quantity
\eqref{probquantity11}, the dependence on dynamics of the system \eqref{dynamics1} given as:

\begin{lemma}\label{lemmaProbabcon}
	Given $N \in \mathbb{N}$ a control policy $\pi = \{ \mu_k \}_{ k=0,\ldots,N-1 } \in U_{N-1}$ and a set $S \subset \textbf{X}$,
	
	\begin{equation}\label{probidentity1}
	P(x_k \in S, k = 0,1,\ldots,N) = \int_S \mathcal{I}(N,x)dx,
	\end{equation}
	
	where for any $x \in \textbf{X}$ and $ k = 1,2,\ldots,N $
	
	\begin{equation}\label{defI}
	\mathcal{I}(k,x) = \begin{cases}
	I_S(x_0), k=0, \\ \int_S p_{f(z,\mu_{k-1})}(x) \mathcal{I}(k-1,z) dz, k = 1,2,\ldots, N
	\end{cases}
	\end{equation}
\end{lemma}

\textit{Proof:} By induction. For $k=0$,

\begin{equation}
P(x_0 \in S) =  \begin{cases}
1,  \ \    if \ x_0 \in S\\ 0, \ otherwise
\end{cases}
\end{equation}

Suppose that \eqref{probidentity1} holds for step $k-1$. By Bayes formula,
the following chain of equalities holds:

\begin{gather}
P(x_0, x_1, \dots, x_k \in S ) = \int_S P(x_k \in S \mid x_{k-1} \in dx,  x_{k-2} \in S, \ldots) \nonumber \\ \cdot P(x_{k-1} \in dx ,x_{k-2} \in S, \dots, x_0 \in S)  = \int_S (\int_S p_{f(x,u_{k-1})}(z) dz ) \mathcal{I}(k-1,x) dx = \nonumber \\
\int_S ( \int_S p_{f(x,\mu_{k-1})}(z) \mathcal{I}(k-1,x) dx) dz = \int_S \mathcal{I}(k,z)dz.
\end{gather}

Thus, \eqref{probidentity1} holds for step $k$.

Inspired by \cite{article}, we now define the cost function associated
to the probability quantity \eqref{probquantity11} and hence to the
Stochastic Invariance Problem. We introduce the following cost function
$V$ which associates a real number $V(k,x, \pi^k) \in [0,1]$ to a triple $(k, x, \pi^k)$ by:

\begin{equation} {\small \label{costV}
	V(k,x, \pi^k) = \begin{cases}
	I_S(x), \ \ k=N; \\ \int_S V(k+1,z,\pi^{k+1}) p_{f(x, \mu_k)}(z) dz,  \quad k = 0,1,\ldots, N-1
	\end{cases} }
\end{equation}

All control policies $\mu $ are such that the cost
function $V$ as in \eqref{costV}, is well–defined, due to properties explained in section \ref{GP}. The following result
establishes a formal functional relationship between $V$ and $\mathcal{I}$ and hence between $V$ and the probability quantity \eqref{probquantity11}.

\begin{proposition}\label{IVidentityprop}
	Given $N \in \mathbb{N} $ a control policy $ \pi \in U_{N-1}$ and a safe set $S$, for any $k=0,1,\ldots,N$:
	\begin{equation}\label{IVidentity}
	\int_S \mathcal{I}(N,x)dx = \int_S V(k,x, \pi^k) \mathcal{I}(k,x)dx.
	\end{equation}
\end{proposition}

\textit{Proof:} By induction. By definition of $\mathcal{I}$ in \eqref{defI},

\begin{equation}
\int_S \mathcal{I}(N,x)dx = \int_S I_S(x) \mathcal{I}(N,x)dx = \int_S V(N,x, \pi^N) \mathcal{I}(N,x)dx.
\end{equation}

Hence, the statement holds for $ k = N $. By proceeding
backwards, we suppose that \eqref{IVidentity} is true for step $k$ and we
prove that \eqref{IVidentity} is true for step $k-1$. By replacing Equation \eqref{defI} into Equation \eqref{IVidentity}, and by Equation \eqref{costV}, we have:

\begin{gather}
\int_S \mathcal{I}(N,x)dx = \int_S V(k,x, \pi^k) ( \int_S p_{f(x,\mu_{k-1})}(x) \mathcal{I}(k-1,z)dz ) dx \nonumber \\
= \int_S(  \int_S ( V(k,x, \pi^k) p_{f(z,\mu_{k-1})}(x) dx ) \mathcal{I}(k-1,z))dz \nonumber 
= \int_S V(k-1,x, \pi^{k-1}) \mathcal{I}(k-1,z) dz,
\end{gather}

and hence the result follows. 

The result above gives the way for rewriting the probability
quantity \eqref{probquantity11} in terms of the cost function $V$ :

\begin{proposition}\label{proposVprob}
	Given $N \in \mathbb{N} $ a control policy $ \pi \in U_{N-1}$ and a safe set $S$,
	\begin{equation}
	P(x_k \in S, \forall k = 0,\ldots,N) = \int_S V( 0,x, \pi) I_S(x) dx.
	\end{equation}
\end{proposition}

\textit{Proof:} By applying Proposition \eqref{IVidentityprop} at step $k = 0$, and
by Lemma \ref{lemmaProbabcon}, the statement holds. 

By Proposition \ref{proposVprob}, it is possible to rewrite the Stochastic
Invariance Problem, as follows.

\begin{problem}\label{problem2}
	Given a finite time horizon $N \in \mathbb{N} $ and a safe set $S \subset \textbf{X} $, compute:
	\begin{equation}\label{pioptimal}
	\pi^* = \arg\sup_{\pi} \int_S V(0,x,\pi) I_S(x) dx
	\end{equation}
\end{problem}

Problem \ref{problem2}, as reformulated above, 
highlights connections between the Stochastic Invariance Problem and optimal control problems. 
In accordance with Problem \ref{problem11} we formulate optimal control problem for stochastic systems with integrated safety approach as Problem \ref{problem3} from section \ref{RA}. Similarly to Problem \ref{problem2} we redefine the optimal control problem, just substituting a safe set $S$ for $X_T$ in the last step. Before giving the main proof, we need the following technical result.

\begin{lemma}\label{lemma5}
	Given a measurable subset $S$ of $\textbf{X}$, let be
	
	\begin{gather}
	\eta_1 :  \textbf{X} \times \textbf{U} \rightarrow \mathbb{R}, \nonumber \\ 
	\eta_2 :  \textbf{X}   \rightarrow \mathbb{R} \nonumber,
	\end{gather}
	such that $\eta_2(x) \geq 0$ for any $x \in S$ and let $U_0$ be the class of feedback functions,
	i.e. $U_0 = \{ \mu : \textbf{X} \rightarrow \textbf{U} \}$. The optimal control policy $\mu^* \in U_0 $, solving the following optimization problem:
	\begin{equation}\label{12}
	\sup_{\mu \ \in \ U_0} \int_S \eta_1(x, \mu(x)) \eta_2(x) dx
	\end{equation}
	is such that for any $\mu \in U_0$,
	\begin{equation}\label{13}
	\eta_1(x,\mu(x)) \leq \eta_1(x,\mu^*(x))
	\end{equation}
	almost everywhere with respect to $x \in S$ Conversely, if $ \mu^* \in U_0 $ satisfies \eqref{13} for any $x \in S$, then $\mu^*$ is the
	solution to the optimization problem \eqref{12}.
\end{lemma}

\textit{Proof:} For the sake of contradiction, suppose that
there exists a measurable set $A \subset S $, with non–zero
Lebesgue measure, such that for any $x \in A$, inequality
\eqref{13} is not true. Then, there exists a feedback control policy $\bar{\mu} \in U_0$ such that

\begin{equation}
\int_A \eta_1(x,\bar{\mu}(x))\eta_2(x) dx  > \int_A \eta_1(x,\mu^*(x)) \eta_2(x) dx. \nonumber
\end{equation}

Define the following control policy,

\begin{equation}
\hat{\mu} = \begin{cases} 
\mu^*(x), if x \in S\setminus A, \\ 
\bar{\mu}(x), if x \in A. \nonumber
\end{cases}
\end{equation}

Then,

\begin{gather}
\int_{x \in S} \eta_1(x,\hat{\mu}(x))\eta_2(x) dx = \int_{S\setminus A} \eta_1(x,\mu^*(x))\eta_2(x) dx + \int_A \eta_1(x,\bar{\mu}(x))\eta_2(x) dx \nonumber \\
>  \int_{S\setminus A} \eta_1(x,\mu^*(x))\eta_2(x) dx + \int_A \eta_1(x,\mu^*(x))\eta_2(x) dx 
=  \int_S \eta_1(x,\mu^*(x))\eta_2(x) dx,  \nonumber
\end{gather}

and hence $\mu^*$ is not the optimal control policy that solves
problem \eqref{12}. The second part of the statement is trivial. 

\subsection{Proof of Theorem \ref{mainthm1}}\label{AppendixB}

\textit{Proof:} For $k = 0,1, \ldots, N-2$, let $J_k^*(x)$ be the optimal cost for the $(N - k)$-stage problem 
that starts at state $x$ and time $k$, and ends at time $N$, i.e. $J_k^* (x) = \sup_{\pi^k} V'(k,x,\pi^k)$. 
For $k = N$, we define $J^*_N (x) = I_{X_T}(x)$  
Now we will show by induction that the functions $J^*_k(\cdot)$ are
equal to the functions $J_k(\cdot)$, as defined in \eqref{recurcive1}, so that for
$k = 0$ we obtain the desired result. Assume that for some
$k$ and all $x$, we have that $J^*_{k+1}(x) = J_{k+1}(x)$. Then, since
$\pi^k = (\mu_k, \pi^{k+1})$, we have for all $x$,

\begin{gather}
J^*_k (x) = \sup_{(\mu_k,\pi^{k+1})} V'(k,x,\pi^k) = \sup_{(\mu_k,\pi^{k+1})} \int_S V'(k+1,z,\pi^{k+1}) p_{f(x,\mu_{k})}(z)dz \nonumber \\
= \sup_{\mu_k} \int_S (\sup_{\pi^{k+1}} V'(k+1,z,\pi^{k+1})) p_{f(x,\mu_{k})}(z)dz 
= \sup_{\mu_k} \int_S J^*_{k+1}(z)  p_{f(x,\mu_{k})}(z) dz  \nonumber \\
= \sup_{\mu_k} \int_S J_{k+1}(z)  p_{f(x,\mu_{k})}(z) dz 
= \sup_{u_k \in \textbf{U}} \int_S J_{k+1}(z)  p_{f(x,\mu_{k})}(z) dz = J_{k}(x), \nonumber
\end{gather}

completing the induction. The second equality holds by
definition of $V'$ in \eqref{costV}. In the third equality, we moved the
supremum over $\pi^{k+1}$ inside the integral because of Lemma 
\ref{lemma5} and of the principle of optimality argument ( see e.g. \cite{bertsekas2005dynamic} ). 
In the forth equality, we used the definition of $J^*_{k+1}(x)$,
and in the fifth equality, we used the induction hypothesis.
Finally, in the sixth equality, we converted the supremum
over $\mu_k$ to a supremum over $u_k$, using the fact that for any
function $F$ of $z$ and $u$, we have:

\begin{equation}
\sup_{\mu \in U_1} F(z,\mu(z)) =  \sup_{u \in \textbf{U}} F(z,u).  \nonumber
\end{equation}

\printbibliography

\end{document}